%% file: template.tex
\documentclass{Interspeech}

\interspeechcameraready

\title{\ours: A Multimodal Deep Fusion Multi-Stage Training Framework for Speech Emotion Recognition in Naturalistic Conditions}

\author[affiliation={1}]{Georgios}{Chatzichristodoulou$^*$}
\author[affiliation={1,2}]{Despoina}{Kosmopoulou$^*$}
\author[affiliation={1}]{Antonios}{Kritikos$^*$}
\author[affiliation={1}]{Anastasia}{Poulopoulou}
\author[affiliation={1,3}]{Efthymios}{Georgiou$^\dagger$}
\author[affiliation={3}]{Athanasios}{Katsamanis}
\author[affiliation={3}]{Vassilis}{Katsouros}
\author[affiliation={1,2}]{Alexandros}{Potamianos}

\affiliation{ECE}{National Technical University of Athens}{Greece}
\affiliation{Archimedes}{Athena RC}{Greece}
\affiliation{ILSP}{Athena RC}{Greece}

\email{\{el20125, el20001, el20025, el20074, efthygeo\}@mail.ntua.gr, \\ \hspace{.2cm} \{nkatsam, vsk\}@athenarc.gr, potam@central.ntua.gr} 

\keywords{speech emotion recognition, deep fusion, meta-classifier, multi-stage training}

\usepackage{comment}
\usepackage{algorithm}
\usepackage{algpseudocode}
\usepackage{multicol}
\usepackage{multirow}
\usepackage{hyperref}
\usepackage{float}
\usepackage{pgfplots}
\pgfplotsset{compat=1.17}
\usepackage{algorithmicx}
\usepackage{amsmath}
\usepackage{hyperref}
\usepackage{url}

\usepackage{xcolor,colortbl}

\hypersetup{
    colorlinks=true,
    linkcolor=blue,
    filecolor=magenta,      
    urlcolor=cyan
}

\input{tex/abbrev}

\input{tex/defn}

\begin{document}

\maketitle

\def\thefootnote{*}\footnotetext{Equal contribution authors}\def\thefootnote{\arabic{footnote}}
\def\thefootnote{$\dagger$}\footnotetext{Corresponding authors}
\def\thefootnote{$\ddagger$}

\begin{abstract}
SER is a challenging task due to the subjective nature of human emotions and their uneven representation under naturalistic conditions. We propose \ours, a multimodal framework with a four-stage training pipeline, which effectively handles class imbalance and emotion ambiguity. The first two stages train an ensemble of classifiers that utilize DeepSER, a novel extension of a deep cross-modal transformer fusion mechanism from pretrained self-supervised acoustic and linguistic representations. Manifold MixUp is employed for further regularization. The last two stages optimize a trainable meta-classifier that combines the ensemble predictions. Our training approach incorporates human annotation scores as soft targets, coupled with balanced data sampling and multitask learning. \ours ranked 1\textsuperscript{st} in Task 1: Categorical Emotion Recognition in the Interspeech 2025: Speech Emotion Recognition in Naturalistic Conditions Challenge.
\end{abstract}

\input{tex/intro_1}
\input{tex/rel_work}

\input{tex/method_1}
\input{tex/exp_setup}
\input{tex/exp_results_1}
\input{tex/disc}
\input{tex/ackn}

\bibliographystyle{IEEEtran}
\bibliography{mybib}

\end{document}

%% file: tex/abbrev.tex
\newcommand{\alert}[1]{{\color{red}{#1}}}

\newcommand{\Th}[1]{\textsc{#1}}

\newcommand{\tb}[1]{\textbf{#1}}

\newcommand{\citeme}[1]{\alert{[X]}}
\newcommand{\refme}[1]{\alert{(X)}}

\makeatletter
\newcommand*\bdot{\mathpalette\bdot@{.7}}
\newcommand*\bdot@[2]{\mathbin{\vcenter{\hbox{\scalebox{#2}{$\m@th#1\bullet$}}}}}
\makeatother

\makeatletter
\DeclareRobustCommand\onedot{\futurelet\@let@token\@onedot}
\def\@onedot{\ifx\@let@token.\else.\null\fi\xspace}

\def\eg{\emph{e.g}\onedot} 
\def\ie{\emph{i.e}\onedot}

\makeatother

%% file: tex/defn.tex
\newcommand{\relu}{\operatorname{ReLU}}
\newcommand{\tflayer}{\operatorname{TransformerLayer}}
\newcommand{\mixup}{\operatorname{MixUp}}
\newcommand{\pool}{\operatorname{POOL}}
\newcommand{\tfenc}{\operatorname{ENC}}

\newcommand{\ours}{\textsc{Medusa}\xspace}

\definecolor{Gray0}{gray}{0.4}
\definecolor{Gray1}{gray}{0.75}
\definecolor{Gray2}{gray}{0.80}
\definecolor{Gray2}{gray}{0.82}
\definecolor{Gray3}{RGB}{205, 246, 250} 
\definecolor{Gray4}{gray}{0.95}
\definecolor{my_Green}{RGB}{0,140,0}

\definecolor{LightBlue1}{RGB}{173, 216, 230} 
\definecolor{LightBlue2}{RGB}{135, 206, 250} 
\definecolor{LightBlue3}{RGB}{176, 224, 230} 

\definecolor{CB91_Blue}{HTML}{2CBDFE}
\definecolor{CB91_Green}{HTML}{47DBCD}
\definecolor{CB91_Pink}{HTML}{F3A0F2}
\definecolor{CB91_Purple}{HTML}{9D2EC5}
\definecolor{CB91_Violet}{HTML}{661D98}
\definecolor{CB91_Amber}{HTML}{F5B14C}

\definecolor{Point_Color}{HTML}{9b59b6}
\definecolor{Hull_Line}{HTML}{34495e}
\definecolor{Combination}{HTML}{e67e22}

%% file: tex/intro_1.tex
\section{Introduction}
Speech Emotion Recognition (SER) is a fundamental affective computing problem that aims to automatically identify emotional states from human speech~\cite{911197}, with significant technological applications in healthcare~\cite{app11114782}, intelligent driving~\cite{10.1007/978-3-030-21074-8_32}, and call centers~\cite{app10134653}. Despite the advancement in deep learning (DL) methodologies~\cite{emotion2vec2024, 7472669}, SER faces several critical challenges. The primary is the limited availability of large annotated SER databases across domains and languages. Moreover, the inherent ambiguity in human emotion perception leads to annotator disagreement, and in naturalistic conditions the speech emotion palette is heavily imbalanced, making category-specific classification and data collection difficult. These challenges require careful consideration in real-world SER systems.

The \emph{Interspeech 2025: Speech Emotion Recognition in Naturalistic Conditions Challenge}~\cite{Naini_2025} offers such a realistic SER scenario exhibiting both annotator disagreement (ambiguity) and class-imbalance across emotion categories. Utilizing the \emph{MSP-Podcast} corpus, a collection of speech segments from audio-sharing websites~\cite{msppodcast}, participants are required to develop SER systems for either categorical emotion recognition or emotional attribute prediction. The first task involves classification across eight distinct emotional classes (anger, happiness, sadness, fear, surprise, contempt, disgust and neutral), while the second predicts emotional attributes, \ie, arousal (calm to active), valence (negative to positive), and dominance (weak to strong), within a range of 1 to 7. 

In this work, we introduce \ours, a multimodal and multi-stage training framework for Task 1: ``Categorical Emotion Recognition''. \ours incorporates DeepSER, a deep fusion architecture, which effectively captures acoustic and linguistic information, through a four-stage training pipeline. Stage 1 trains the DeepSER model on the complete dataset, while Stage 2 resumes training with balanced subsets, to address class imbalance. 
Stage 3 trains a meta-classifier to combine ensemble predictions, and Stage 4 implements a model soup~\cite{wortsman2022modelsoupsaveragingweights} over different meta-classifiers. \ours achieves first place in Task 1 of the challenge. Our contributions are:
\begin{itemize}
    \item We develop DeepSER, a novel extension of a deep fusion architecture~\cite{georgiou19_interspeech}, which integrates transformer-based models and expands beyond two modalities for enhanced cross-modal integration.
    \item We present a two-stage training recipe for DeepSER, utilizing detailed annotator scores and Manifold MixUp regularization, to address annotation ambiguity and class-imbalance.
    \item We use a trainable meta-classifier over the ensemble, which learns model and class reliability coefficients (Stage 3),  and is further enhanced via model soup averaging (Stage 4).
    \item The overall framework, \ours, achieves 1\textsuperscript{st} place in Task 1 of {\it Interspeech 2025: Speech Emotion Recognition in Naturalistic Conditions Challenge}.
    \footnote{Code is available at: \href{https://github.com/emopodntua/medusa}{https://github.com/emopodntua/medusa}}
\end{itemize}



%% file: tex/rel_work.tex
\section{Related Work}

SER spans a wide range of setups, from acted emotions~\cite{Busso2008} to naturalistic~\cite{msppodcast} and in-the-wild scenarios~\cite{mosei}. Research approaches are typically supervised and use raw speech signals, mel-spectrograms and extracted features~\cite{ser_survey_2021}. Other learning approaches employ self-supervised learning (SSL), as shown by emotion2vec~\cite{emotion2vec2024} and WavLLM~\cite{wavllm2024}. Multimodal methods integrate acoustic, linguistic and visual signals. \ours focuses specifically on acoustic and linguistic modalities. Our fusion strategy, DeepSER, draws inspiration from DHF~\cite{georgiou19_interspeech} and extends it. Several components of \ours are based on conceptually related work in computational paralinguistics: soft target learning strategy~\cite{sof_targets_2018}, mixing-based regularization~\cite{verma2019manifold, Georgiou2024} and mulitask learning~\cite{cai21b_interspeech}. For ensemble methods, model diversity~\cite{chen20241st} is a helpful factor. The meta-classifier~\cite{cieliebak-etal-2014-meta} idea has also been investigated independently. \ours combines these approaches into a unified SER framework.

%% file: tex/method_1.tex
\section{Method}
\subsection{Overview}
The novelty of our approach stems from three key directions: 1) architecture, 2) data utilization and 3) training recipe. The proposed architecture, \emph{DeepSER}, is a novel extension of a deep fusion approach~\cite{georgiou19_interspeech}. Different DeepSER instances are combined via a meta-classifier and form \ours. For data utilization, \ours leverages detailed human annotator scores as soft targets, \eg, two annotators voting for happy and one for neutral.

\begin{figure}
    \centering
    \includegraphics[width=1\linewidth]{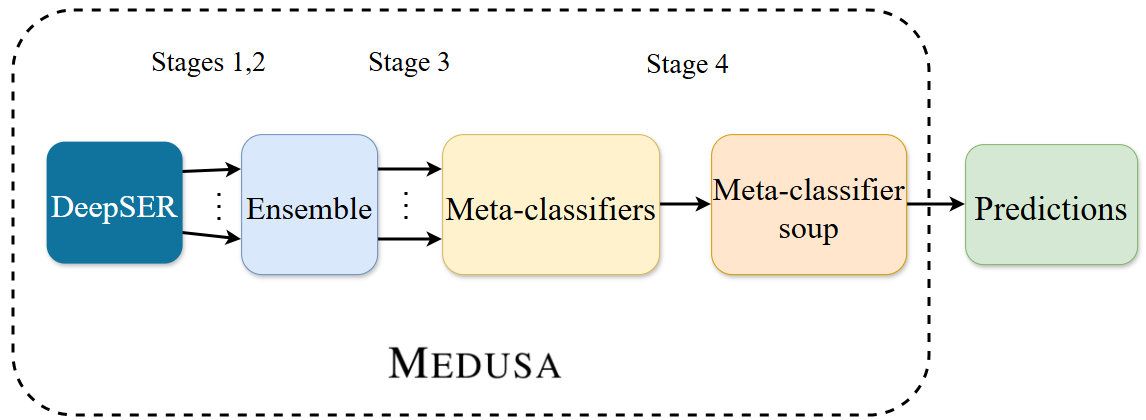}
    \caption{An overview of \ours. 
    Different modalities are combined on each model using DeepSER. In total, 14 different such models are trained leveraging 2 or 3 modalities as their backbone. The training process employs standard training (Stage 1) on the full training dataset, balanced subset training (Stage 2) and a linear meta-classifier on top for ensembling (Stage 3). Six different runs of the meta-classifier are merged into a model soup by averaging the weights (Stage 4).}
    \label{fig:over}
\end{figure}

The training recipe follows a four-staged pipeline. The first two stages train the DeepSER models with soft targets and Manifold MixUp~\cite{verma2019manifold} (M.MixUp) regularization, with Stage 1 using standard training, and Stage 2 employing balanced subset training. 
The final two stages of the pipeline form the \ours ensemble, where a meta-classifier learns to combine the posterior distributions of individual classifiers using Stage's 2 balanced training approach. The synergy of these components yields significant performance improvements. An overview of \ours is illustrated in Figure \ref{fig:over}.

\subsection{The MSP-Podcast dataset}
The MSP-Podcast~\cite{msppodcast} Interspeech 2025 SER Challenge version~\cite{Naini_2025} consists of audio recordings, their transcriptions, their respective forced alignments, speaker metadata (IDs and genders), and detailed annotator decisions for both categorical and regression tasks. For the held-out test set, only audio files are included. \ours utilizes the audio files, detailed voting decisions (soft targets) and emotional attribute ratings (multi-task).

\noindent
\textbf{Soft Targets:} Understanding the nature of annotator (dis)agreement provides crucial insights, \ie, whether disagreement stems from poor sample quality or proximity to multiple emotions. Using the full decision vector, rather than one-hot representations, offers two advantages: it enables us to leverage the entire dataset, including samples without consensus ($\sim$20K samples), and helps the model develop more \textit{nuanced} decision boundaries. We normalize these decision vectors consistently across all samples.

\noindent \textbf{Emotional Attributes:} The emotional attributes (valence, arousal, dominance) are incorporated into the model's multitask learning objective, providing additional regularization.

\subsection{Feature Extraction}
We leverage multiple pretrained self-supervised learning (SSL) models to achieve representational pluralism. 

\noindent
\textbf{Speech Features:}\renewcommand{\thefootnote}{\fnsymbol{footnote}}
We extract features using transformer-based~\cite{NIPS2017_attention} models: 
\texttt{WavLM}\footnote[1]{\href{https://huggingface.co/microsoft/wavlm-large}{https://huggingface.co/microsoft/wavlm-large}}~\cite{Chen_2022} and 
\texttt{HuBERT}\footnote[2]{\href{https://huggingface.co/facebook/hubert-large-ls960-ft}{https://huggingface.co/facebook/hubert-large-ls960-ft}}~\cite{hsu2021hubertselfsupervisedspeechrepresentation} which process raw audio signals, 
and the \texttt{Whisper-v3}\textsuperscript{$\ddagger$} encoder~\cite{radford2022robustspeechrecognitionlargescale}, which processes mel-spectrograms.

\noindent
\textbf{Text Features:}
Linguistic information is extracted using the \texttt{Whisper-v3}\footnote[3]{\href{https://huggingface.co/openai/whisper-large-v3}{https://huggingface.co/openai/whisper-large-v3}} Automatic Speech Recognition (ASR) model, which transcribes speech into text that is then fed into \texttt{RoBERTa}\footnote[4]{\href{https://huggingface.co/FacebookAI/roberta-large}{https://huggingface.co/FacebookAI/roberta-large}}~\cite{liu2019robertarobustlyoptimizedbert} and \texttt{ModernBERT}\footnote[5]{\href{https://huggingface.co/answerdotai/ModernBERT-large}{https://huggingface.co/answerdotai/ModernBERT-large}}~\cite{warner2024smarterbetterfasterlonger} language encoders for feature extraction. ASR is used consistently across training and testing to prevent train/test distribution shift.

\renewcommand{\thefootnote}{\fnsymbol{footnote}}

\subsection{Deep fusion architecture - DeepSER}
Drawing inspiration from the Deep Hierarchical Fusion (DHF) scheme~\cite{georgiou19_interspeech}, which combines unimodal representations at multiple layers, we propose a novel extension named DeepSER.
Our extension makes two key advancements. First, we replace the original LSTM-based~\cite{lstm} architecture with transformers~\cite{NIPS2017_attention} to better capture unimodal and cross-modal relationships. Second, we extend the fusion mechanism to handle $N$ arbitrary modalities, enabling integration of multiple heterogeneous feature streams.

DeepSER utilizes a base transformer encoder architecture (Algorithm \ref{alg:enc}) as its fundamental building block for both unimodal and fusion processing. The overall fusion process (Algorithm \ref{alg:deep}) operates by first extracting intermediate representations from each unimodal encoder's layers. These representations are then fed into fusion encoders, which combine them hierarchically, \ie, each fusion stage processes both the current unimodal features and, when available, the fusion representations from previous stages, enabling deep multimodal integration.
\ours's implementation utilizes DeepSER architecture with a two-layer transformer encoder (\Th{Enc}), and for 2 and 3 modalities\footnote[6]{Our implementation, shown here for a two-layer transformer with three modalities, can be easily extended to any number of layers and modalities. A detailed technical report covering these extensions will be available in our GitHub repository.}.

\begin{algorithm}[!htb]
\caption{Base Transformer Encoder (\Th{Enc})}
\footnotesize
\begin{algorithmic}[0]
\State\hspace*{-1.25em} \textbf{Defn:} $W_i(\cdot)$: linear map, $\sigma(\cdot): \relu$ 
\Require Tensor $\mathbf{x}$: (B,L,D)
\Ensure List of (hidden) tensors $\mathbf{h}_i$

\State $\mathbf{x} \gets W_2(\sigma(W_1(\mathbf{x})))$ \Comment{map seq. to fixed dimension}
\State $\mathbf{h}_1 \gets \tflayer(\mathbf{x})$ \Comment{First Layer: (B,L,D)}
\State $\mathbf{h}_2 \gets \tflayer(\mathbf{h}_1)$ \Comment{Second Layer: (B,L,D)}
\State $\mathbf{h}_3 \gets \pool(\mathbf{h}_2)$ \Comment{Pooling Layer: (B,D)}
\State \Return $[\mathbf{h}_1, \mathbf{h}_2, \mathbf{h}_3]$

\end{algorithmic}
\label{alg:enc}
\end{algorithm}

\begin{algorithm}[!htb]
\caption{DeepSER}
\footnotesize
\begin{algorithmic}
\Require Tensors $\mathbf{h}_0, \mathbf{g}_0, \mathbf{z}_0$
\Require Labels $\bf{y}_c, \bf{y}_r$
\Ensure Output tensors $\bf{y}_c, \bf{y}_r$

\State $[\bf{h}_1, \bf{h}_2, \bf{h}_3] \gets \textcolor{CB91_Violet}{\tfenc}_{h}(\bf{h}_0)$  \Comment{$\bf{h}$ \textcolor{CB91_Violet}{unimodal} enc.}
\State $[\bf{g}_1, \bf{g}_2, \bf{g}_3] \gets \textcolor{CB91_Violet}{\tfenc}_{g}(\bf{g}_0)$  \Comment{$\bf{g}$ \textcolor{CB91_Violet}{unimodal} enc.}
\State $[\bf{z}_1, \bf{z}_2, \bf{z}_3] \gets \textcolor{CB91_Violet}{\tfenc}_{z}(\bf{z}_0)$  \Comment{$\bf{z}$ \textcolor{CB91_Violet}{unimodal} enc.}
\State $[\_ , \bf{f}^{(1)}_2, \_ ] \gets \textcolor{my_Green}{\tfenc}_{f^{(1)}}(\bf{h}_1 || \bf{g}_1 || \bf{z}_1)$  \Comment{$\bf{f}^{(1)}$ 1st \textcolor{my_Green}{fusion}}
\State $[\_ , \_ , \bf{f}^{(2)}_3 ] \gets \textcolor{my_Green}{\tfenc}_{f^{(2)}}(\bf{h}_2 || \bf{g}_2 || \bf{z}_2 || \bf{f}_2^{(1)})$  \Comment{$\bf{f}^{(2)}$ 2nd \textcolor{my_Green}{fusion}}
\State $\bf{x} \gets W(\bf{h}_3 || \bf{g}_3 || \bf{z}_3 || \bf{f}^{(2)}_3)$  \Comment{pooled representation fusion}
\State $\bf{x}, \bf{y}_c, \bf{y}_r \gets \mixup(\bf{x}, \bf{y}_c, \bf{y}_r)$  \Comment{M.MixUp}
\State $\bf{y}_c \gets W_c(\sigma(\bf{x}))$  \Comment{Classification Map}
\State $\bf{y}_r \gets W_r(\sigma(\bf{x}))$  \Comment{Regression Map}
\State \Return $\mathbf{y}_c, \mathbf{y}_r$
\end{algorithmic}
 \label{alg:deep}
\end{algorithm}

\subsection{Meta-classifier}
\ours leverages multiple SSL models for both text and speech, leading to several (DeepSER) trained models with different configurations. To effectively combine these models' complementary strengths, we implement a \emph{meta-classifier} layer. While traditional ensemble methods like soft or hard voting proved ineffective to combine diverse posterior predictions, we found that a simple linear layer trained on top of the models' predictions yields superior results. This meta-classifier learns to combine individual model predictions and also, through balanced training, helps mitigate the class imbalance problem by promoting more uniform distribution in \ours's output.

\subsection{Multistage Training Recipe}    
To address the significant class imbalance problem, we introduce a four-stage training process for \ours.
During \textbf{Stage 1}, we train a multimodal DeepSER instance (model) on the full training dataset. 
At \textbf{Stage 2}, we continue training using balanced data subsets, where each epoch uses randomly sampled balanced subsets from the training data, ensuring equal (sample) representation across emotion categories. Stages 1 and 2 are repeated for each classifier in the ensemble. 
At \textbf{Stage 3}, we train a linear meta-classifier to combine the posteriors of the DeepSER ensemble, using the balanced sampling of Stage 2.
At \textbf{Stage 4} (post-training),  we enhance meta-classifier's robustness via model soup~\cite{wortsman2022modelsoupsaveragingweights} weight averaging across different seeds.

\subsection{Manifold MixUp} 

Manifold MixUp~\cite{verma2019manifold} (M.MixUp) is a regularization technique that interpolates between latent representations of input examples. For a pair of examples with hidden representations $\textbf{h}_i, \textbf{h}_j \in \mathbb{R}^n$ and corresponding labels $\textbf{y}_i, \textbf{y}_j$, M.MixUp computes mixed features $\tilde{\textbf{h}} = \lambda \textbf{h}_i + (1-\lambda) \textbf{h}_j$ and labels $\tilde{\textbf{y}} = \lambda \textbf{y}_i + (1-\lambda) \textbf{y}_j$, where $\lambda \in [0,1]$ is drawn from a $\text{Beta}(\alpha, \alpha)$ distribution. Drawing connections to vicinal risk minimization (VRM)~\cite{NIPS2000_vrm}, this approach creates virtual training points in the latent space, proving particularly beneficial for our SER task with its class imbalance and annotator ambiguity challenges. In our implementation, M.MixUp is applied with probability $p$ during training, before the classification head of DeepSER.

\subsection{Loss Function}
The training objective for Stages 1, 2, 3 is defined as 
\begin{equation}
\mathcal{L} = \lambda_1 \cdot \mathcal{L}_{CE} + \lambda_2 \cdot \mathcal{L}_{MSE}
\label{eq:loss_function}
\end{equation}
where $\mathcal{L}_{CE}$ is the weighted cross-entropy loss for emotion classification task, and $\mathcal{L}_{MSE}$ is the mean squared error (MSE) loss for the emotional attribute prediction. This multitask scheme serves as additional regularization. 
During Stage 1, we address class imbalance by weighting the cross-entropy loss with $\alpha_c = (\frac{N}{|C| \cdot f_c})^{0.5}$, where $N$ is the total samples, $|C|$ is the number of classes, and $f_c$ is the frequency of class $c$. Stages 2, 3 use uniform class weights due to their balanced sampling strategy.

%% file: tex/exp_setup.tex
\section{Experimental Setup}
DeepSER models are trained with batch size 16 and learning rate \texttt{1e-5} during Stages 1 \& 2, with M.MixUp's $p=0.3$. The Stage 3 meta-classifier uses batch size 128 and learning rate \texttt{1e-3} without M.MixUp. DeepSER utilizes transformer encoders (see Algorithm ~\autoref{alg:enc}) with two layers across both unimodal and fusion processing, with hidden dimension of 1024.
The loss weights are set to $\lambda_1 = 1.5$ and $\lambda_2 = 0.4$. We optimize \ours's parameters across all Stages using AdamW~\cite{loshchilov2019decoupledweightdecayregularization}. 
We create an in-house $90$-$10$ train-test split, used consistently across all models and stages. Model selection is based on the best average Macro-F1 ($F1$)\footnote{Throughout this paper, $F1$ refers to Macro-F1 score.} on balanced internal held-out test-sets. All models are implemented in PyTorch~\cite{paszke2019pytorchimperativestylehighperformance} and trained on one RTX 3090 GPU.

%% file: tex/exp_results_1.tex
\section{Experimental Results}
\subsection{Challenge Result} 
\textbf{\ours ranked 1\textsuperscript{st}} on the Challenge's leaderboard, demonstrating remarkably consistent performance across all metrics on the held-out test set: Macro-F1 of \textbf{0.4316}, Micro-F1 of \textbf{0.4319}, and Accuracy of \textbf{0.4319}. This balance between macro and micro metrics suggests that \ours effectively handles both majority and minority emotion classes, addressing the class imbalance challenge, which is the primary problem of the Challenge. The configurations of the $14$ individual DeepSER models utilized by \ours are illustrated in~\autoref{tab:ensemble}.

\begin{table}[h]
    \centering    
    \footnotesize
    \caption{Configurations of \ours's DeepSER models and their $(F1)$ score in our internal test-set. \Th{Id}: refers to DeepSER model ID; \Th{Model}: refers to SSL feature extractor; \Th{F1}: macro-F1 score.  
    }
    \begin{tabular}{ccccc}
        \toprule
        \Th{Id} & \Th{Model} 1  & \Th{Model} 2 & \Th{Model} 3 & \Th{F1}$\uparrow$ \\
        \midrule 
        1 & HuBERT & RoBERTa & - & 0.368 \\
        \rowcolor{Gray3}
        2 & WavLM & HuBERT & RoBERTa & \tb{0.434} \\
        3 & WavLM & Whisper & HuBERT & 0.394 \\
        4 & WavLM & ModernBERT &  -  & \tb{0.427} \\
        5 & WavLM & RoBERTa &   -   & \tb{0.430} \\
        6 & WavLM & RoBERTa &   -   & 0.425 \\
        7 & WavLM & RoBERTa & - & \tb{0.437} \\
        \rowcolor{Gray3}
        8 & WavLM & RoBERTa & - & \tb{0.440} \\
        \rowcolor{Gray3}
        9 & Whisper & ModernBERT & - & 0.406 \\
        10 & Whisper & ModernBERT & - & 0.403 \\
        11 & Whisper & RoBERTa & - & 0.392 \\
        12 & Whisper & RoBERTa & - & 0.405 \\
        \rowcolor{Gray3}
        13 & Whisper & WavLM & - & 0.403 \\
        14 & Whisper & WavLM & ModernBert & 0.410 \\
        \midrule
         & Average DeepSER &  &  & 0.413 \\
         & Top DeepSER &  &  & 0.440 \\
         \rowcolor{LightBlue3}
         & \tb{\ours} & & & \tb{0.472} \\
        \bottomrule
    \end{tabular}
    
    \label{tab:ensemble}
\end{table}

\begin{table*}[t!]
    \centering
    \footnotesize
    \caption{DeepSER-level ablation. \Th{F1}: denotes macro-F1 score; \Th{Model} ID: denotes a model from Table 1; \Th{Acc}: denotes accuracy; \Th{Avg}: denotes average performance across the examined models; $\Delta$: relative \Th{F1} degradation (minus sign) from DeepSER.}
   \begin{tabular}{l *{8}{c} c c c }
        \toprule
        \multirow{2}{*}{\Th{}} & \multicolumn{2}{c}{\Th{Model 2}} & \multicolumn{2}{c}{\Th{Model 8}} & \multicolumn{2}{c}{\Th{Model 9}} & \multicolumn{2}{c}{\Th{Model 13}} & \multicolumn{2}{c}{\Th{Avg}} & \Th{$\Delta$ (\%)} \\
        \cmidrule(lr){2-3} \cmidrule(lr){4-5} \cmidrule(lr){6-7} \cmidrule(lr){8-9} \cmidrule(lr){10-11}
        & \Th{F1} & \Th{Acc} & \Th{F1} & \Th{Acc} & \Th{F1} & \Th{Acc} & \Th{F1} & \Th{Acc} & \Th{F1} & \Th{Acc} & \\
        \midrule
         \rowcolor{LightBlue3}
         DeepSER & 0.434 & 0.441 & 0.440 & 0.450 & 0.406 & 0.421 & 0.403 & 0.419 & 0.421 & 0.432 & -- \\
         w. One-hot targets & 0.431 & 0.440 & 0.402 & 0.423 & 0.368 & 0.392 & 0.353 & 0.384 & 0.389 & 0.410 & -8.23 \\
         w. Late Fusion & 0.413 & 0.428 & 0.410 & 0.428 & 0.393 & 0.405 & 0.368 & 0.395 & 0.396 & 0.414 & -6.31 \\
         w/o MixUp & 0.434 & 0.444 & 0.421 & 0.437 & 0.395 & 0.409 & 0.384 & 0.406 & 0.409 & 0.424 & --2.93 \\
         w/o multitask & 0.436 & 0.440 & 0.411 & 0.429 & 0.410 & 0.437 & 0.390 & 0.409 & 0.412 & 0.429 & -2.18 \\
         w/o Stage 2 & 0.434 & 0.440 & 0.419 & 0.432 & 0.403 & 0.450 & 0.409 & 0.418 & 0.416 & 0.435 & -1.20 \\
        \bottomrule
    \end{tabular}
    \label{tab:ablation}
\end{table*}


\begin{table*}[htbp]
    \centering
    \footnotesize
    \caption{\ours-level ablation: We combine different number of models from Table 1 within the \ours framework and experiment with different ensemble aggregation techniques. \Th{F1}: denotes macro-F1 score; \Th{Acc}: denotes accuracy.}
   \begin{tabular}{lccccccc}
        \toprule
        \multirow{2}{*}{\Th{Technique}} & \multicolumn{2}{c}{\ours(\Th{14 Models})} & \multicolumn{2}{c}{\ours(\Th{Top-7 Models})} & \multicolumn{2}{c}{\ours(\Th{Top-5 Models})}\\
\cmidrule(lr){2-3} \cmidrule(lr){4-5} \cmidrule(lr){6-7} 
        & \Th{F1} & \Th{Acc} & \Th{F1} & \Th{Acc}  & \Th{F1} & \Th{Acc}   \\
        \midrule 
        Majority Voting & 0.439 & 0.454 & 0.436 & 0.449 & 0.437 & 0.450 \\
        Soft Voting & 0.439 & 0.453 & 0.439 & 0.453 & 0.439 & 0.453  \\
        \rowcolor{Gray3}
        Meta-classifier (Single) & 0.471 & 0.470 & 0.465 & 0.465 & 0.468 & 0.462 \\
        \rowcolor{LightBlue3}
        Meta-classifier (Soup) & 0.472 & 0.471 & 0.467 & 0.466 & 0.445 & 0.441 \\
        \bottomrule
    \end{tabular}
    
    \label{tab:ablation_ens}
\end{table*}

\subsection{DeepSER-level Ablation}
\noindent
\textbf{Overview:}
We conduct a DeepSER-level ablation study to investigate the impact of each architectural and training component. We select models $2$, $8$, $9$ and $13$ from~\autoref{tab:ensemble} for this analysis, as they represent both top-performing configurations and diverse model architectures with varying feature sets and SSL extractors. 
In each experiment, we replace a key component with its standard implementation and evaluate on our balanced test sets. The ablated components include: deep fusion replaced with late fusion, soft targets with one-hot labels, and training without (w/o) M.MixUp, w/o Stage 2 (balanced training), and w/o multitask loss.

\noindent
\textbf{Results:}
All components consistently improve F1 performance, with results averaged across multiple DeepSER models for robust conclusions. The complete ablation results are presented in~\autoref{tab:ablation}. Soft targets emerge as the most impactful component, as removing them leads to an average 8.23\% relative degradation in F1 score. This suggests that leveraging detailed human annotations helps models better capture nuanced emotion boundaries. Deep fusion proves vital for performance, with its removal causing a 6.31\% degradation, highlighting the importance of deep multimodal information integration for SER. M.MixUp's removal also results in significant performance drops, likely due to losing the virtual training points that smooth decision boundaries around existing soft targets. Removing Stage 2 (balanced training) and multitask learning also degrades performance. These results demonstrate the effectiveness and synergistic nature of our architectural and training components.

\subsection{\ours-level Ablation}
\noindent
\textbf{Overview:}
We conduct a \ours-level ablation study exploring two key directions: 1) the optimal number of models in the ensemble and 2) the effectiveness of different ensemble aggregation techniques. We compare four voting approaches: hard voting, soft voting, single-run meta-classifier and meta-classifier soup, where we average meta-classifier weights over multiple runs to smooth decision boundaries~\cite{wortsman2022modelsoupsaveragingweights}.

\noindent
\textbf{Results:}
The results, presented in~\autoref{tab:ablation_ens}, demonstrate that both meta-classifier variants consistently outperform soft and hard voting approaches, indicating the significance of this approach. Compared to the top DeepSER model (see~\autoref{tab:ensemble}), \ours shows a 7.23\% relative improvement. Furthermore, \autoref{tab:ablation_ens} shows that while meta-classifier (single run) performs better when combining the top-5 models of~\autoref{tab:ensemble}, the meta-classifier soup shows slight advantages when incorporating more models, \ie, 7 and 14. Based on these findings, we opted for the meta-classifier soup in our final \ours submission. Finally, we limit our ensemble to 14 models, as doubling from 7 to 14 models showed diminishing improvements. 

%% file: tex/disc.tex
\section{Discussion}

In this paper, we present \ours, a multimodal and multi-stage framework that achieved first place in the Interspeech 2025 Challenge. The \ours framework integrates DeepSER, a novel deep fusion architectural extension, with soft targets, mixing-based regularization, and multi-stage training to address annotation ambiguity and class imbalance. A meta-classifier (soup) layer enables robust ensemble prediction aggregation. Our ablation studies reveal the significant impact of each component: for \ours ensemble, the meta-classifier layer and model diversity are the most important factors, while for DeepSER, soft targets, deep fusion architecture, and mixing-based regularization emerge as the most crucial elements for addressing SER in naturalistic conditions.

Future research directions include integrating forced alignment and speaker information to enrich the feature space, and exploring modality-agnostic augmentations for improved generalization. The deep fusion architecture could be extended to incorporate more SSL models and feature types, while the meta-classifier could be enhanced by utilizing pre-logit representations and deeper architectures. These extensions could further improve the framework's ability to handle the complexities of emotion recognition in naturalistic conditions.

%% file: tex/ackn.tex
\section{Acknowledgments}

This work is supported by the European Union’s Horizon Europe Research and Innovation Programme under Grant No 101061303, project PREMIERE (Performing arts in a new era: AI and VR tools for better understanding, preservation, enjoyment and accessibility).
This work has been partially supported by project MIS 5154714 of the National Recovery and Resilience Plan Greece 2.0 funded by the European Union under the NextGenerationEU Program. 